\documentclass{article}
\usepackage{ijcai19}

\usepackage{times}
\usepackage{soul}
\usepackage{url}
\usepackage[hidelinks]{hyperref}
\usepackage[utf8]{inputenc}
\usepackage[small]{caption}
\usepackage{graphicx}
\usepackage{amsmath}
\usepackage{booktabs}
\usepackage{algorithm}
\usepackage{algorithmic}
\usepackage{booktabs} % For formal tables
\usepackage{helvet}
\usepackage{courier}
\usepackage{textcomp}
\usepackage{xcolor}
\usepackage{booktabs}
\usepackage{multirow}
\usepackage{color}
\usepackage{subfigure}
\usepackage{threeparttable}
\usepackage{tabularx}
\usepackage{hyperref}
\usepackage{amsthm}
\usepackage{balance}
\title{Attributed Graph Clustering: A Deep Attentional Embedding Approach}

\author{
Chun Wang$^1$\and
Shirui Pan$^2$\and
Ruiqi Hu$^1$\and
Guodong Long$^1$\and
Jing Jiang$^1$\And
Chengqi Zhang$^1$\\
\affiliations
$^1$Centre for Artificial Intelligence, University of Technology Sydney, Australia\\
$^2$Faculty of IT, Monash University, Australia\\
\emails
\{chun.wang-1, ruiqi.hu\}@student.uts.edu.au,
shirui.pan@monash.edu,\\
\{guodong.long, jing.jiang, chengqi.zhang\}@uts.edu.au
}

\begin{document}
\maketitle
\begin{abstract}
Graph clustering is a fundamental task which discovers communities or groups in networks. Recent studies have mostly focused on developing deep learning approaches to learn a compact graph embedding, upon which classic clustering methods like $k$-means or spectral clustering algorithms are applied. These two-step frameworks are difficult to manipulate and usually lead to suboptimal performance, mainly because the graph embedding is not goal-directed, i.e., designed for the specific clustering task. In this paper, we propose a goal-directed deep learning approach, Deep Attentional Embedded Graph Clustering (\texttt{DAEGC} for short). Our method focuses on attributed graphs to sufficiently explore the two sides of information in graphs. By employing an attention network to capture the importance of the neighboring nodes to a target node, our DAEGC algorithm encodes the topological structure and node content in a graph to a compact representation, on which an inner product decoder is trained to reconstruct the graph structure. Furthermore, soft labels from the graph embedding itself are generated to supervise a self-training graph clustering process, which iteratively refines the clustering results. The self-training process is jointly learned and optimized with the graph embedding in a unified framework, to mutually benefit both components. Experimental results compared with state-of-the-art algorithms demonstrate the superiority of our method.

%In recent years, the relatively mainstream research idea on this topic has advanced to exploit the interplay between graph structure and node content information with some deep learning based models, and learn graph embedding that could be later applied to classic clustering methods like $k$-means or spectral clustering. This kind of two-steps frameworks are harder to manipulate and could not achieve a global optimal solution. Furthermore, graph clustering is naturally unsupervised problem with no label guidance, which further increase the difficulty. In this paper, we proposed a self-training graph attentional clustering method. We integrate graph structure and node content information into a attentional autoencoder for embedding learning. Unlike the common unsupervised approaches, we generate fake labels from the graph embedding itself to supervise the training process. This self-training clustering component is further optimized together with the autoencoder in an end-to-end manner, jointly learn graph embedding and perform clustering at the same time. Experimental results compared with 12 state-of-the-art algorithms demonstrate the superiority of our proposed method.
\end{abstract}

%\keywords{Grpah clustering, deep learning, attention}

\section{Introduction}

The development of networked applications has resulted in an overwhelming number of scenarios in which data is naturally represented in graph format rather than flat-table or vector format. Graph-based representation characterizes individual properties through node attributes, and at the same time captures the pairwise relationship through the graph structure. Many real-world tasks, such as the analysis of citation networks, social networks, and protein-protein interaction, all rely on graph-data mining skills. However, the complexity of graph structure has imposed significant challenges on these graph-related  learning tasks, including graph clustering, which is one of the most popular topics.

Graph clustering aims to partition the nodes in the graph into disjoint groups.  %\cite{DBLP:conf/aaai/BojchevskiG18,chen2017revisiting}. 
Typical applications include community detection \cite{hastings2006community}, %DBLP:conf/aaai/LiSHZ18
group segmentation \cite{kim2006customer}, and functional group discovery in enterprise social networks \cite{hu2016co}. Further for attributed graph clustering, a key problem is how to capture the structural relationship %between nodes 
and exploit the node content information.

To solve this problem, more recent studies have resorted to deep learning techniques to learn compact representation to exploit the rich information of both the content and structure data \cite{wu2019comprehensive}.
Based on the learned graph embedding,  simple clustering algorithms such as $k$-means are applied. Autoencoder is a mainstream solution for this kind of embedding-based approach \cite{cao2016deep,tian2014learning}, as the autoencoder based hidden representation learning approach can be applied to purely unsupervised environments. %Many autoencoder based graph clustering algorithms already exist. 
% \citet{tian2014learning} considered the similarity of autoencoder and spectral clustering and learned a latent representation for clustering through sparse autoencoder. \citet{cao2016deep} proposed a deep graph representation model for clustering by capturing structure information through random surfing. The recently developed GAE and VGAE \cite{kipf2016variational} based on graph convolutional network (GCN) can also be adopted for graph clustering analysis.

Nevertheless, all these embedding-based methods are two-step approaches. The drawback is that the learned embedding may not be the best fit for the subsequent graph clustering task, and the graph clustering task is not beneficial to the graph embedding learning. To achieve mutual benefit for these two steps, a goal-directed training framework is highly desirable. However, traditional goal-directed training models are mostly applied to the classification task. For instance, \cite{kipf2016variational} proposed graph convolutional networks for networked data. Fewer studies on goal-directed embedding methods for graph clustering exist, to the best of our knowledge.  %An end to end model for deep graph clustering is highly desired.

%Task oriented embedding, such as semi-supervised GCN has demonstrated impressive results on node classification tasks. However, for clustering, to the best of our knowledge, less studies exist. 

\begin{figure}
\vspace{1mm}
\centering
\includegraphics[width=0.99\linewidth]{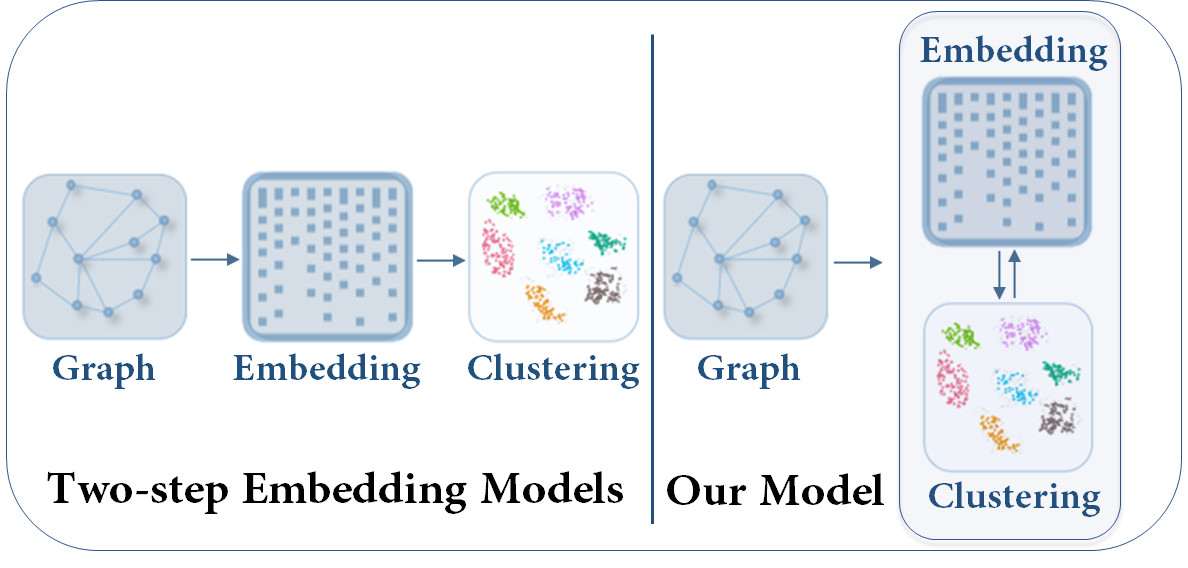}
\caption{The difference between two-step embedding learning models and our model
}
\label{fig:endtoend}
\vspace{-2mm}
\end{figure}

% \textbf{Our Approach}
Motivated by the above observations, we propose a goal-directed graph attentional autoencoder based attributed graph clustering framework in this paper. To exploit the interrelationship of various-typed graph data, %we turn to graph attention networks (GAT). 
we develop a graph attentional autoencoder to learn latent representation%, which integrates both content and structure information
. The encoder exploits both graph structure and node content with a graph attention network, and multiple layers of encoders are stacked to build a deep architecture for embedding learning. The decoder on the other side, reconstruct the topological graph information and manipulates the latent graph representation. %Nevertheless, such autoencoder based framework is naturally unsupervised which will no doubt weaken the performance compared with supervised models. Therefore, 
We further employ a self-training module, which takes the ``\textit{confident}" clustering assignments as soft labels to guide the optimizing procedure. By forcing the current clustering distribution approaching a hypothetical better distribution, in contrast to the two-step embedding learning-based methods (shown in Fig \ref{fig:endtoend}), this specialized clustering component simultaneously learns the embedding and performs clustering in a unified framework, thereby achieving better clustering performance. Our contributions can be summarized as follows:

\begin{itemize}
\item We develop the first graph attention-based autoencoder to effectively integrate both structure and content information for deep latent representation learning.

\item We propose a new goal-directed framework for attributed graph clustering. The framework 
%self-training process to manipulate the embedding learning of the graph autoencoder, which 
jointly optimizes the embedding learning and graph clustering, to the mutual benefit of both components. %make it possible to jointly learn embedding and perform clustering in an end-to-end manner.

\item The experimental results show that our algorithm outperforms state-of-the-art graph clustering methods.
\end{itemize}

\section{Related Work}

%Our work is closely related to graph clustering, graph attentional neural networks and self-training autoencoder based deep clustering algorithms. We briefly review some of these works in this section.

\subsection{Graph Clustering} 
Graph clustering has been a long-standing research topic. 
Early methods have taken various shallow approaches to graph clustering. \cite{girvan2002community} used centrality indices to find community boundaries and detect social communities. \cite{hastings2006community} applied belief propagation to community detection and determined the most likely arrangement of communities.  %\citet{newman2006finding} %newman2006modularity
%computed the eigenvectors of the graph Laplacian to perform clustering. 
Many embedding learning based approaches apply an existing clustering algorithm on the learned embedding \cite{wang2017community}. 
To handle both content and structure information, %probabilistic model \cite{cohn2001missing}, 
relational topic models \cite{sun2009itopicmodel,chang2009relational}, co-clustering method \cite{guo2019cfond}, %chang2009relational
and content propagation \cite{liu2015community} have also been widely used.  %Dam and Velden for instance, collect profile information and apply an MCA $k$-means algorithm to cluster Facebook users \cite{van2015online}.

The limitations of these methods are that (1) they only capture either parts of the network information or shallow relationships between the content and structure data, and (2) they are directly applied on sparse original graphs. As a result, these methods cannot effectively exploit the graph structure or the interplay between the graph structure and the node content information.

In recent years, benefiting from the development of deep learning, graph clustering has progressed significantly. Many deep graph clustering algorithms employ autoencoders, adopting either the variational autoencoder \cite{kipf2016variational}, sparse autoencoder \cite{tian2014learning,hu2017deep}, adversarially regularized method \cite{pan2019learning} or  denoising autoencoder \cite{cao2016deep} to learn deep representation for clustering. However, these methods are two-step methods, whereas the algorithm presented in this paper is a unified approach.

\begin{figure*}[h]
\centering
\includegraphics[width=0.85\linewidth]{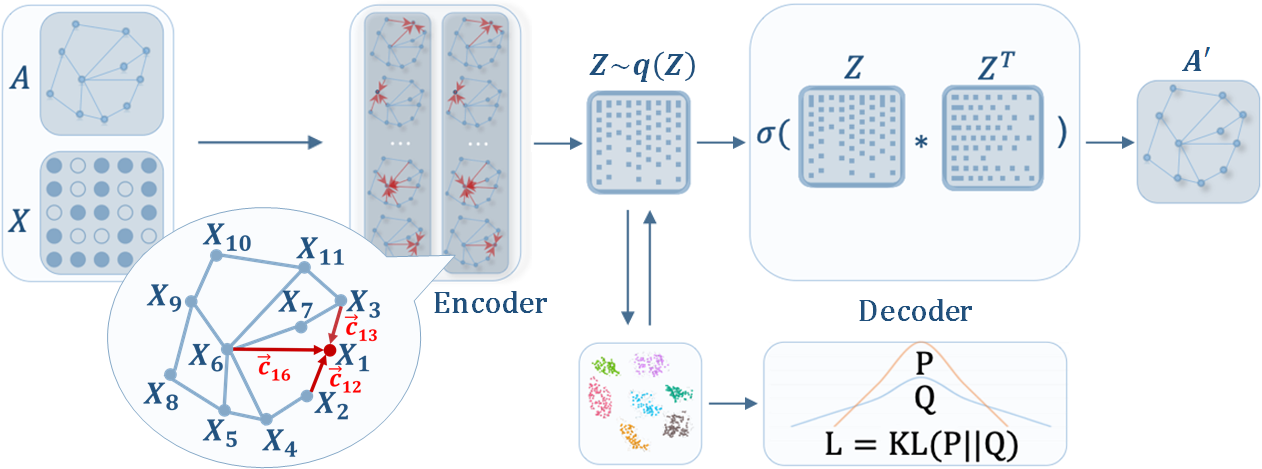}
\caption{\small The conceptual framework of Deep Attentional Embedded Graph Clustering (DAEGC). Given a graph $G=(V, E, X)$, DAEGC learns a hidden representation $Z$ through a graph attention-based autoencoder, and manipulates it with a self-training clustering module, which is optimized together with the autoencoder and perform clustering during training.
}
\label{fig:framework}
\end{figure*}

\subsection{Deep Clustering Algorithms} 
Autoencoders have been a widely used tool in the deep learning area, especially for unsupervised learning tasks such as clustering \cite{wang2017mgae} and anomaly detection \cite{zhou2017anomaly}. %Autoencoders basically consist of an encoder mapping the input feature $X$ to some hidden representation $h(X)$ and a decoder mapping it back to reconstruct the input feature. The parameters of the autoencoder can be learned through minimizing the reconstruction error.

Deep Embedded Clustering (DEC) is a specialized clustering technique %that was  proposed by %Xie et al.
\cite{xie2016unsupervised}. This method employs a stacked denoising autoencoder learning approach. After obtaining the hidden representation of the autoencoder by pre-train, %rather than minimizing the reconstruction error through a decoder, 
the encoder pathway is fine-tuned by a defined Kullback-Leibler divergence clustering loss. %DEC perform $k$-means on the obtained hidden embedding of the autoencoder and acquire an initial clustering of the original image data. Rather than assign each data point to a certain cluster with the highest probability, it produce a distribution which reserce the probability of every possible assignment. Then DEC defined a new distribution, which is a transformation of the former distribution and put more emphasis on data points assigned with high confidence. The clustering loss is defined as the distance between the two distributions and the distribution is forced optimizing with the minimizing of the clustering loss.
\cite{guo2017improved} considered that the defined clustering loss could corrupt the feature space and lead to non-representative features, so they added back the decoder and optimized the reconstruction error together with the clustering loss. %In this way, an end-to-end clustering algorithm is built jointly learning embedded features and performing clustering with local structure preservation.

There have since then been increasing algorithms based on such deep clustering framework \cite{dizaji2017deep,guo2017deep}. %guo2017deep
However, as far as we know, they are only designed for data with flat-table representation. For graph data, complex structure and content information need to be carefully exploited, and goal-directed clustering for graph data is still an open problem in this area. 
%only few work adopt such deep clustering to graph based analysis. Yang et al. combine deep clustering with influence propagation and inject the structure information to the hidden embedding of the autoencoder for clustering loss optimizing \cite{yang2017graph}, which is the only related work on graph data to the best of our knowledge. Compared with their algorithm, we argue that our approach combined graph structure into the autoencoder and could capture the most salient information of not only the node attributes but also the graph topological structure, and thereby make use of deep clustering in a more efficient way.

\section{Problem Definition and Overall Framework}

We consider clustering task on attributed graphs in this paper. A graph is represented as $G = (V,E,X)$, where $V=\{v_i\}_{i=1,\cdots,n} $ consists of a set of nodes, $E=\{e_{ij}\}$ is a set of edges between nodes. The topological structure of graph $G$ can be represented by an adjacency matrix $A$, where $A_{i,j} = 1$ if $(v_{i},v_{j})\in E$; otherwise $A_{i,j} = 0$. $X = \{x_1;\dots ;x_n\} $ are the attribute values where $x_i \in R^{m}$ is a real-value attribute vector associated with vertex $v_i$.

Given the graph $G$, graph clustering aims to partition the nodes in $G$ into $k$ disjoint groups $\{G_1, G_2, \cdots, G_k\}$, so that nodes within the same cluster are generally: (1) close to each other in terms of graph structure while distant otherwise; and (2) more likely to have similar attribute values.

% \vspace{1mm}\noindent\textbf{Overall Framework}
%In this paper, we construct a graph attention network to solve this problem. 
Our framework is shown in Fig \ref{fig:framework} and consists of two parts: a graph attentional autoencoder and a self-training clustering module.
\begin{itemize}
    \item \textbf{Graph Attentional Autoencoder:} Our autoencoder takes the attribute values and graph structure as input, and learns the latent embedding by minimizing the reconstruction loss.

    \item \textbf{Self-training Clustering:} The self-training module performs clustering based on the learned representation, and in return, manipulates the latent representation according to the current clustering result.
\end{itemize}

 We jointly learn the graph embedding and perform clustering in a unified framework, so that each component benefits the other.

\section{Proposed Method}
In this section, we present our proposed Deep Attentional Embedded Graph Clustering  (\texttt{DAEGC}). We first develop a graph attentional autoencoder which effectively integrates both structure and content information to learn a latent representation. Based on the representation, a self-training module is proposed to guide the clustering algorithm towards better performance. 

\subsection{Graph Attentional Autoencoder}
%The Graph Attentional Autoencoder consists of two parts, the graph attentional encoder and the link prediction decoder.

\subsubsection{Graph attentional encoder:} To represent both graph structure ${A}$ and node content ${X}$ in a unified framework, we develop a variant of the graph attention network \cite{velickovic2017graph} as a graph encoder.
%is developed from the graph attention networks (GATs) which were formerly used in semi-supervised classification tasks. 
The idea is to learn hidden representations of each node by attending over its neighbors, to combine the attribute values with the graph structure in the latent representation. The most straightforward strategy to attend the neighbors of a node is to integrate its representation equally with all its neighbors. However, in order to measure the importance of various neighbors, different weights are given to the neighbor representations in our layer-wise graph attention strategy:

\begin{equation}
z_i^{l+1} = \sigma (\sum_{j \in N_i} \alpha_{ij}Wz_j^{l}).
\label{eq:zil+1}
\end{equation}

Here, $z_i^{l+1}$ denotes the output representation of node $i$, and $N_i$ denotes the neighbors of $i$. $\alpha_{ij}$ is the attention coefficient that indicates the importance of neighbor node $j$ to node $i$, and $\sigma$ is a nonlinerity function.
To calculate the attention coefficient $\alpha_{ij}$, we measure the importance of neighbor node $j$ from both the aspects of the attribute value and the topological distance. 

From the perspective of attribute values, the attention coefficient $\alpha_{ij}$ can be represented as a single-layer feedforward neural network on the concatenation of $x_i$ and $x_j$ with weight vector $\overrightarrow{a} \in R^{2m'}$:
\begin{equation}
c_{ij} = \overrightarrow{a}^T[Wx_i||Wx_j].
\label{eq:cij}
\end{equation}

Topologically, neighbor nodes contribute to the representation of a target node through edges. GAT considers only the 1-hop neighboring nodes (first-order) for graph attention \cite{velickovic2017graph}. As graphs have complex structure relationships, we propose to exploit high-order neighbors in our encoder. We obtain a proximity matrix by considering $t$-order neighbor nodes in the graph:
%closer to node $i$ are naturally expected to be given larger weights. Therefore we add an extra weight for each neighbor to value more the nearer (e.g. first order) neighbors:
\begin{equation}
M = (B + B^2 + \cdots + B^t)/t, 
\label{eq:M}
\end{equation}
here $B$ is the transition matrix where $B_{ij} = 1/d_i$ if $e_{ij} \in E$ and $B_{ij} = 0$ otherwise. $d_i$ is the degree of node $i$. Therefore $M_{ij}$ denotes the topological relevance of node $j$ to node $i$ up to $t$ orders. In this case, $N_i$ means the neighboring nodes of $i$ in $M$. i.e., $j$ is a neighbor of $i$ if   $M_{ij}>0$. $t$ could be chosen flexibly for different datasets to balance the precision and efficiency of the model.  

The attention coefficients are usually normalized across all neighborhoods $j \in N_i$ with a softmax function to make them easily comparable across nodes:
\begin{equation}
\alpha_{ij} = \textrm{softmax}_j(c_{ij}) = \frac{\textrm{exp}(c_{ij})}{\sum_{r \in N_i}\textrm{exp}(c_{ir})}.
\label{eq:alphaij1}
\end{equation}

Adding the topological weights $M$ and an activation function $\delta$ (here $\textrm{LeakyReLU}$ is used), the coefficients can be expressed as:
\begin{equation}
\alpha_{ij} = \frac{\textrm{exp}(\delta M_{ij}(\overrightarrow{a}^T[Wx_i||Wx_j]))}{\sum_{r \in N_i}\textrm{exp}(\delta M_{ir}(\overrightarrow{a}^T[Wx_i||Wx_r]))}.
\label{eq:alphaij2}
\end{equation}

We have $x_i = z_i^{0}$ as the input for our problem, and stack two graph attention layers:
\begin{equation}
z_i^{(1)} =  \sigma (\sum_{j \in N_i} \alpha_{ij}W^{(0)}x_j),
\label{eq:zi1}
\end{equation}
\begin{equation}
z_i^{(2)} =  \sigma (\sum_{j \in N_i} \alpha_{ij}W^{(1)}z_j^{(1)}),
\label{eq:zi2}
\end{equation}
in this way, our encoder encodes both the  structure and the node attributes into a hidden representation, i.e., we will have $z_i = z_i^{(2)}$.
% \begin{equation}
% Z = \prod_{i=1}^n z_i = \prod_{i=1}^n z_i^{(2)}.
% \end{equation}

\vspace{.2cm}
\subsubsection{Inner product decoder:}
There are various kinds of decoders, which reconstruct either the graph structure, the attribute value, or both. As our latent embedding already contains both content and structure information, we choose to adopt a simple inner product decoder to predict the links between nodes, which would be efficient and flexible:
%In our method, we choose to reconstruct the graph structure, as our algorithm will be more flexible and will thus fit situations in which no content information is available. We use a simple inner product decoder which predicts whether there is a link between two nodes. The reconstructed link prediction layer is trained based on the hidden graph representation:
% \begin{equation}
% \textrm{Recon} = \prod_{i=1}^n \prod_{j=1}^n D(\hat{A}_{ij}|z_i, z_j),  \textrm{with}
% \end{equation}
% \begin{equation}
% D(\hat{A} _{ij}=1 | z_i, z_j) = \textrm{sigmoid}({z_i}^\top z_j),
% \end{equation}
% where $\hat{A}$ is the reconstructed adjacency matrix $A$ of the graph.

\begin{equation}
\hat{A}_{ij} = \textrm{sigmoid}({z_i}^\top z_j),
\label{eq:Ahatij}
\end{equation}
where $\hat{A}$ is the reconstructed structure matrix of the graph.

\vspace{.2cm}
\subsubsection{Reconstruction loss:}
We minimize the reconstruction error  by measuring the difference between  $A$ and $\hat A$:
\begin{equation}
L_r = \sum_{i=1}^n loss(A_{i,j}, \hat{A}_{ij}).
\label{eq:Lr}
\end{equation}
%In our paper, the binary cross-entropy loss function is used as the reconstruction loss.

\subsection{Self-optimizing Embedding}

One of the main challenges for graph clustering methods is the nonexistence of label guidance. The graph clustering task is naturally unsupervised and feedback during training as to whether the learned embedding is well optimized cannot therefore be obtained. To confront this challenge, we develop a self-optimizing embedding algorithm as a solution. %we are inspired by the recently proposed Deep Embedded Clustering (DEC) \cite{xie2016unsupervised} algorithm which give a possible solution.

Apart from optimizing the reconstruction error, we input our hidden embedding into a self-optimizing clustering module which minimizes the following objective:
\begin{equation}
L_c = KL(P||Q) = \sum_i \sum_u p_{iu} log\frac{p_{iu}}{q_{iu}},
\label{eq:Lc}
\end{equation}
% where $q_{iu}$ is the Student's $t$-distribution \cite{maaten2008visualizing} measured similarity between node embedding $z_i$ and cluster center embedding $\mu_u$, which can be seen as a soft clustering assignment of each node:

Where $q_{iu}$ measures the similarity between node embedding $z_i$ and cluster center embedding $\mu_u$. We measure it with a Student's $t$-distribution so that it could handle different scaled clusters and is computationally convenient \cite{maaten2008visualizing}:

\begin{equation}
q_{iu} = \frac{(1 + || z_i - \mu_u ||^2)^{-1}}{\sum_k{(1 + || z_i - \mu_k ||^2)^{-1}}},
\label{eq:qiu}
\end{equation}
it can be seen as a soft clustering assignment distribution of each node. On the other hand, $p_{iu}$ is the target distribution defined as:
\begin{equation}
p_{iu} = \frac{q_{iu}^2/\sum_i q_{iu}}{\sum_k{(q_{ik}^2/\sum_i q_{ik})}}.
\label{eq:piu}
\end{equation}

Soft assignments with high probability (nodes close to the cluster center) are considered to be trustworthy in $Q$. So the target distribution $P$ raises $Q$ to the second power to emphasize the role of those ``confident assignments". The clustering loss then force the current distribution $Q$ to approach the target distribution $P$, so as to set these ``confident assignments" as soft labels to supervise $Q$'s embedding learning.

% Clearly, $q_{iu}$, the Student's $t$-distribution can be regarded as the probability of assigning sample $i$ to cluster $u$, and $p_{iu}$ is an auxiliary distribution computed from $q_{iu}$.

% The main idea of this objective is to let the embedding learn iteratively from its clustering result.  The target distribution $P$ raises $Q$ to the second power to emphasize the role of those ``confident assignments". The clustering loss is then defined as the KL divergence between the current soft assignment distribution $Q$ and the target distribution $P$, to set these ``confident assignments" as soft labels to supervise the embedding learning.

To this end, we first train the autoencoder without the self-optimize clustering part to obtain a meaningful embedding $z$ as described in Eq.(\ref{eq:zi2}). % which can be regarded as a representation of the nodes in the given graph.
Self-optimizing clustering is then performed to improve this embedding. To obtain the soft clustering assignment distributions of all the nodes $Q$ through Eq.(\ref{eq:qiu}), the $k$-means clustering is performed once and for all on the embedding $z$ before training the entire model, to obtain the initial cluster centers $\mu$. 

Then in the following training, the cluster centers $\mu$ are updated together with the embedding $z$ using Stochastic Gradient Descent (SGD) based on the gradients of $L_c$ with respect to $\mu$ and $z$.

%We take this hidden embedding and run the $k$-means algorithm on it. Through the $k$-means algorithm, we could get an initial group of cluster centers $\mu$ and the  corresponding current soft clustering assignment distributions of all the nodes $Q$ through Eq.(9). 
We calculate the target distribution $P$ according to Eq.(\ref{eq:piu}), and the clustering loss $L_c$ according to Eq.(\ref{eq:Lc}).

The target distribution $P$ works as ``ground-truth labels" in the training procedure, but also depends on the current soft assignment $Q$ which updates at every iteration. It would be hazardous to update $P$ at every iteration with $Q$ as the constant change of target would obstruct learning and convergence. To avoid instability in the self-optimizing process, we update $P$ every 5 iterations in our experiment. 

In summary, we minimize the clustering loss to help the autoencoder manipulate the embedding space using the embedding's own characteristics and scatter embedding points to obtain better clustering performance. 

%\subsection{Self-training Graph Attentional Clustering}
\subsection{Joint Embedding and Clustering Optimization}
We jointly optimize the autoencoder embedding and clustering learning, and define our total objective function as:
\begin{equation}
L = L_r + \gamma L_c,
\label{eq:L}
\end{equation}
where $L_r$ and $L_c$ are the reconstruction loss and clustering loss respectively, $\gamma \geq 0$ is a coefficient that controls the balance in between. 
It is worth mentioning that we could gain our clustering result directly from the last optimized $Q$, and the label estimated for node $v_i$ could be obtained as:
\begin{equation}
s_i = \textrm{arg} \max\limits_u q_{iu},
\label{eq:si}
\end{equation}
which is the most likely assignment from the last soft assignment distribution $Q$.

Our method is summarized in Algorithm \ref{alg:1}. Our algorithm has the following advantages:

\begin{itemize}
\item \textbf{Interplay Exploitation.} The graph attention network-based autoencoder efficiently exploits the interplay between both the structure and content information.

\item \textbf{Clustering Specialized Embedding.} The proposed self-training clustering component manipulates the embedding to improve the clustering performance.

\item \textbf{Joint Learning.} The framework jointly optimizes the two parts of the loss functions, learns the embedding and performs clustering in a unified framework. 
\end{itemize}

\begin{table}[t]
\setlength\tabcolsep{5.2pt}
  \begin{small}
  
%   \vspace{-2mm}
    \begin{tabular}{c|ccccc}
	\toprule  
   	Dataset 	& Nodes   	& Features  & Clusters	& Links  	&  Words    	\\
	\midrule
    Cora  		& 2,708  	& 1,433 	& 7 		& 5,429 	& 3,880,564 		\\
    Citeseer 	& 3,327 	& 3,703 	& 6 		& 4,732 	& 12,274,336 		\\
    Pubmed 		& 19,717 	& 500 		& 3 		& 44,338 	& 9,858,500 		\\
	\bottomrule
    \end{tabular}%
  
  \vspace{-1mm}
  \caption{\normalsize Benchmark Graph Datasets}
  \label{tab:datasets}%
  \end{small}
  \vspace{-2mm}
\end{table}%

\begin{algorithm}[h]
\small
\caption{\normalsize{Deep  Attentional  Embedded  Graph  Clustering}}
\normalsize
\label{alg:1}
\begin{algorithmic}
\REQUIRE ~~\\

Graph $G$ with $n$ nodes;
%Constructed attribute matrix $X \in R^{n \times d}$ and adjacency matrix $A \in R^{n \times n}$  of $G$; 
Number of clusters $k$; ~~Number of iterations $Iter$; ~~Target distribution update interval $T$; ~~Clustering Coefficient $\gamma$.

\ENSURE  ~~\\
Final clustering results.
\STATE
\STATE Update the autoencoder by minimizing Eq.(\ref{eq:Lr}) to get the autoencoder hidden embedding $Z$;
\STATE Compute the initial cluster centers $\mu$ based on $Z$;

\FOR {$l = 0$ to $Iter - 1$}
\STATE Calculate soft assignment distribution $Q$ with $Z$ and $\mu$ according to Eq.(\ref{eq:qiu});
\IF {$l \% T == 0$}
\STATE Calculate target distribution $P$ with $Q$ by  Eq.(\ref{eq:piu});
\ENDIF
\STATE Calculate clustering loss $L_c$ according to Eq.(\ref{eq:Lc});
\STATE Update the whole framework by minimizing Eq.(\ref{eq:L});
\ENDFOR
\STATE Get the clustering results with final $Q$ by  Eq.(\ref{eq:si})  
\end{algorithmic}
\vspace{-1mm}
\end{algorithm}
\vspace{-2mm}

%\vspace{-2mm}
\section{Experiments}

\subsection{Benchmark Datasets}
We used three standard citation networks widely-used for assessment of attributed graph analysis in our experiments, summarized in Table \ref{tab:datasets}. Publications in the datasets are categorized by the research sub-fields.%as they could be assigned to different sub-fields.
%All these datasets consist of scientific publications as nodes, citation relationships as edges and unique words in the documents as features. Publications in the datasets are labeled by the research topics.%as they could be assigned to different sub-fields.

% \vspace{-2mm}
\subsection{Baseline Methods}

We compared a total of ten algorithms with our method in our experiments. The graph clustering algorithms include approaches that use only node attributes or network structure information, and also approaches that combine both. Deep representation learning-based graph clustering algorithms were also compared.%We take into consideration both classes and the following baselines are compared with our algorithm.

\subsubsection{Methods Using Structure or Content Only}

\begin{itemize}
\item\textbf{$K$-means} is the basis of many clustering methods. %Many advanced clustering algorithms involve some kind of transformation of $k$-means clustering or use $k$-means on their embeddings. Here we run $k$-means on our original content data as a benchmark.

\item\textbf{Spectral clustering} uses the eigenvalues to perform dimensionality reduction before clustering. %and is widely used in graph clustering. 

%\item\textbf{Big-Clam} \cite{yang2013overlapping} is a non-negative matrix factorization approach for community detection which takes only the network structure into account. 

\item\textbf{GraphEncoder} \cite{tian2014learning} trains a stacked sparse autoencoder to obtain representation.

\item\textbf{DeepWalk} \cite{perozzi2014deepwalk} is a structure-only representation learning method. %It obtains random walks on graphs and then trains the representation through neural networks.

\item\textbf{DNGR} \cite{cao2016deep} uses stacked denoising autoencoders and encodes each vertex into a low dimensional vector representation. %We try to respectively use structure and content information on DNGR.

\item\textbf{M-NMF} \cite{wang2017community} is a Nonnegative Matrix Factorization model targeted at community-preserved embedding.
\end{itemize}

\subsubsection{Methods Using Both Structure and Content}
\begin{itemize}
%\item \textbf{Circles} \cite{leskovec2012learning} is an attributed graph clustering algorithm which represents overlapping hard-membership approaches for graph clustering.

%\item\textbf{RTM} \cite{chang2009relational} is a relational topic model capturing both structure and content information to learn the topic distributions of documents.

\item\textbf{RMSC} \cite{xia2014robust} is a robust multi-view spectral clustering method. %via low-rank and sparsity decomposition.%, recovering a shared low-rank transition probability matrix for clustering with a transition probability matrix from each view.  
We regard structure and content data as two views of information.

\item\textbf{TADW} \cite{yang2015network} %text-associated DeepWalk, which 
regards DeepWalk as a matrix factorization method and adds the features of vertices for representation learning.

\item\textbf{VGAE \& GAE}  \cite{kipf2016variational} combine graph convolutional network with the (variational) autoencoder to learn representations. 

\item\textbf{DAEGC} is our proposed unsupervised deep attentional embedded graph clustering. \end{itemize}

For representation learning algorithms such as DeepWalk, TADW and DNGR which do not specify the clustering algorithm, we learned the representation from these algorithms, and then applied the $k$-means algorithm on their respective representations, but for algorithms like RMSC which require an alternative algorithm as its clustering method, 
we followed their preference and used the specified algorithms. %The best results we got are reported in this paper.

\begin{table}[t]
  \centering
  \setlength\tabcolsep{4.2pt}
%\scriptsize
%\normalsize
\begin{small}
%\small
  
%   \vspace{-2mm}
    \begin{tabular}{c|c|cccc}
	\toprule  
        & Info. & ACC(\textcolor{red}{$\uparrow $})   & NMI(\textcolor{red}{$\uparrow $})    & F-score(\textcolor{red}{$\uparrow $})  & ARI(\textcolor{red}{$\uparrow $})  \\
\midrule
    $K$-means & C & 0.500 & 0.317 & 0.376  & 0.239 \\
    Spectral & S & 0.398 & 0.297 & 0.332 & 0.174 \\
    %Big-Clam & Structure & 0.272 & 0.007 & 0.281  & 0.001 \\
    GraphEncoder & S & 0.301 & 0.059 & 0.230 & 0.046 \\
    DeepWalk & S & 0.529 & 0.384 & 0.435 & 0.291 \\
    DNGR  & S & 0.419 & 0.318 & 0.340 & 0.142 \\
    M-NMF & S & 0.423 & 0.256 & 0.320 & 0.161 \\
\midrule
    %Circles & Both  & 0.607 & 0.404 & 0.469 & 0.362 \\
    %RTM   & Both  & 0.440 & 0.230 & 0.307 & 0.169 \\
    RMSC  & C\&S  & 0.466 & 0.320 & 0.347 & 0.203 \\
    TADW  & C\&S  & 0.536 & 0.366 & 0.401 & 0.240 \\
    GAE  & C\&S  & 0.530 & 0.397 & 0.415 & 0.293 \\
    VGAE  & C\&S  & 0.592 & 0.408 & 0.456 & 0.347 \\
    DAEGC & C\&S  & \textbf{0.704} & \textbf{0.528} & \textbf{0.682} & \textbf{0.496} \\
	\bottomrule
    \end{tabular}%
  \vspace{-1mm}
  \caption{\normalsize Experimental Results on \textbf{Cora} Dataset}
  \label{tab:result_Cora}%
  \end{small}
%   \vspace{-1mm}
\end{table}%

\begin{table}[t]
  \centering
  \setlength\tabcolsep{4.2pt}
\begin{small}
%   \vspace{-2mm}
    \begin{tabular}{c|c|cccc}
\toprule
        & Info. & ACC(\textcolor{red}{$\uparrow $})   & NMI(\textcolor{red}{$\uparrow $})    & F-score(\textcolor{red}{$\uparrow $})   & ARI(\textcolor{red}{$\uparrow $})  \\
\midrule
    $K$-means & C & 0.544 & 0.312 & 0.413 & 0.285 \\
    Spectral & S & 0.308 & 0.090 & 0.257 & 0.082 \\
    %Big-Clam & Structure & 0.250 & 0.036 & 0.288 & 0.007 \\
    GraphEncoder & S & 0.293 & 0.057 & 0.213 & 0.043 \\
    DeepWalk & S & 0.390 & 0.131 & 0.305 & 0.137 \\
    DNGR  & S & 0.326 & 0.180 & 0.300 & 0.043 \\
    M-NMF & S & 0.336 & 0.099 & 0.255 & 0.070 \\
\midrule
    %Circles & Both  & 0.572 & 0.301 & 0.424 & 0.293 \\
    %RTM   & Both  & 0.451 & 0.239 & 0.342 & 0.203 \\
    RMSC  & C\&S  & 0.516 & 0.308 & 0.404 & 0.266 \\
    TADW  & C\&S  & 0.529 & 0.320 & 0.436 & 0.286 \\
    GAE  & C\&S  & 0.380 & 0.174 & 0.297 & 0.141 \\
    VGAE  & C\&S  & 0.392 & 0.163 & 0.278 & 0.101 \\
    DAEGC & C\&S  & \textbf{0.672} & \textbf{0.397} & \textbf{0.636} & \textbf{0.410} \\
\bottomrule
    \end{tabular}%
    \vspace{-1mm}
    \caption{\normalsize Experimental Results on \textbf{Citeseer} Dataset}
  \label{tab:result_Citeseer}%
  \end{small}
  \vspace{-1mm}
\end{table}%

\begin{figure*}[t]
\centering
\subfigure
{
\label{fig:emb1a}
\includegraphics[width=0.38\columnwidth]{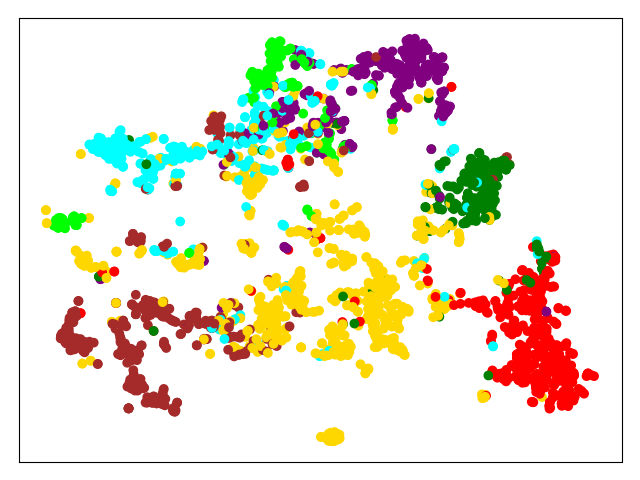}
}
\subfigure
{
\label{fig:emb1b}
\includegraphics[width=0.38\columnwidth]{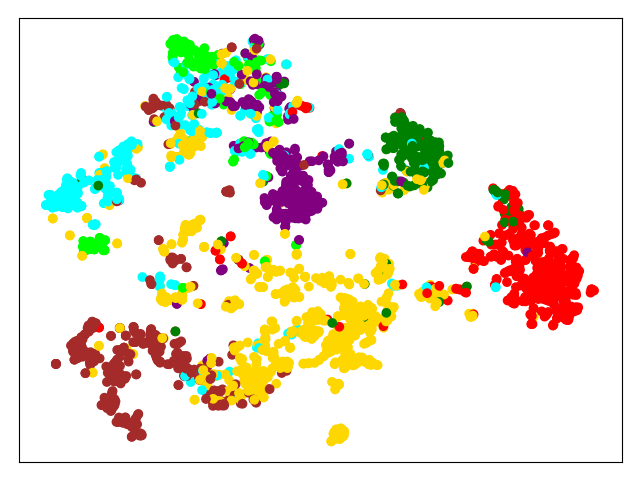}
}
\subfigure
{
\label{fig:emb1c}
\includegraphics[width=0.38\columnwidth]{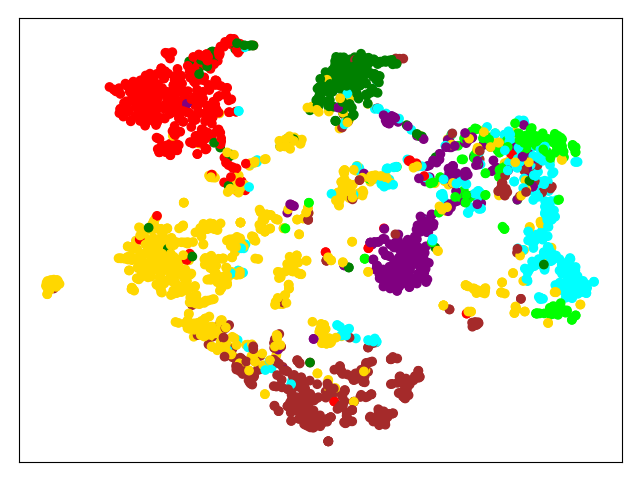}
}
\subfigure
{
\label{fig:emb1d}
\includegraphics[width=0.38\columnwidth]{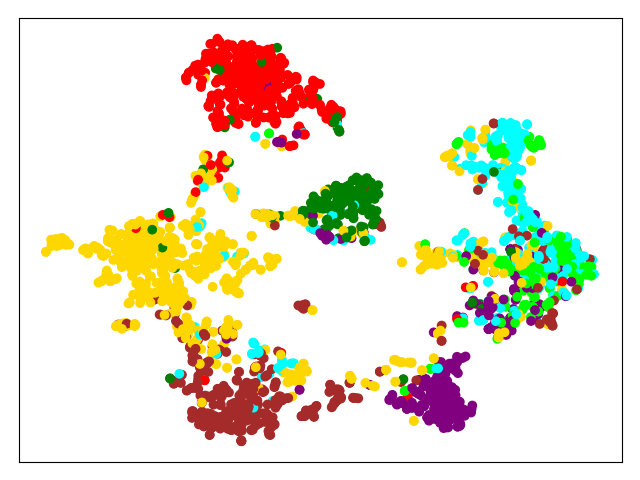}
}
\subfigure
{
\label{fig:emb1e}
\includegraphics[width=0.38\columnwidth]{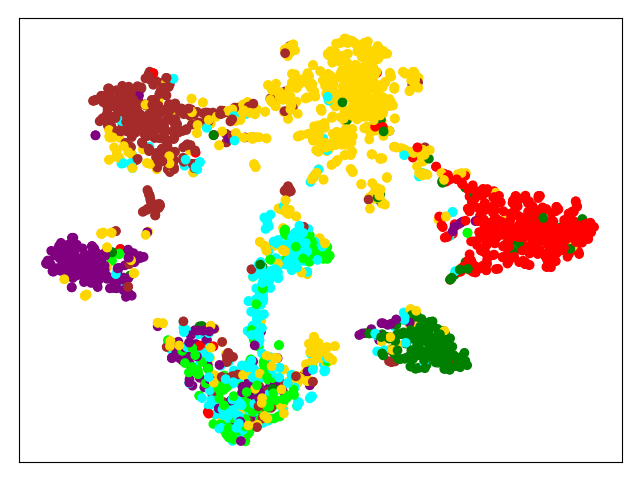}
}
\vspace{-4mm}
\caption{ 2D visualization of the DAEGC algorithm on the Cora dataset during training. The first visualization illustrates the embedding training with the graph attentional autoencoder only, followed by visualizations showing subsequent equal epochs in which the self-training component is included, till the last one being the final embedding visualization.}
\label{fig:tsne}
\end{figure*}

\subsection{Evaluation Metrics \& Parameter Settings}
% Table generated by Excel2LaTeX from sheet 'Citeseer'
\textbf{Metrics:} We use four metrics \cite{xia2014robust} to evaluate the clustering result: Accuracy (ACC), Normalized Mutual Information (NMI), F-score, and Adjusted Rand Index (ARI). A better clustering result should lead to a higher values for all the metrics.

%\begin{itemize}
%	\item \textbf{ACC} is the average performance of label matching clustering results and can be represented as $\sum_i(y_i == f(l_i))/n$, where $f$ is the mapping function which maps category labels to cluster labels. %We need to find the best match for category labels and cluster labels since clustering is an unsupervised process;
%	\item \textbf{NMI} measures the mutual information entropy between the resulting cluster labels and ground truth labels followed by a normalization operation.% which is formulated as:
%	\item \textbf{F-score} is the harmonic mean value of $Precision$ and $Recall$;% : $F = \frac{2P\cdot R}{P+R}$;
%	\item \textbf{Precision} is the fraction of correctly clustered nodes among the retrieved nodes;%: $P = 
%	\item \textbf{Recall} is the fraction of correctly clustered nodes that have been retrieved over the total number of relevant nodes;
%	\item \textbf{AE} $= \sum_{i=1}^k\frac{m_i}{m}e_i$, where $k$ is the cluster number and $m$ is the number of nodes, and $e_i = -\sum_{j=1}^k\frac{m_{ij}}{m_i}log_2\frac{m_{ij}}{m_i}$, with $m_i$ representing the number of nodes in cluster $i$ and $m_{ij}$ representing the number of nodes in cluster $i$ and labeled $j$. 
%	\item \textbf{ARI} is the adjusted rand index ($RI$) that guarantees a value close to 0, where $RI$ measures the percentage of correct clustering decisions;
%
%\end{itemize}

\vspace{.1cm}
\noindent\textbf{Baseline Settings:} For the baseline algorithms, we carefully select the parameters for each algorithm, following the procedures in the original papers. In TADW, for instance, we set the dimension of the factorized matrix to 80 and the regularization parameter to 0.2; % For the DNGR algorithm, we build a three-layers denoising autoencoder with the number of nodes set as 512 and 256 in the hidden layers; 
For the RMSC algorithm, we regard graph structure and node content as two different views of the data and construct a Gaussian kernel on them. We run the $k$-means algorithm 50 times to get an average score for all embedding learning methods for fair comparison. 

\noindent\textbf{Parameter Settings:} For our method, we set the clustering coefficient $\gamma$ to 10. We consider second-order neighbors and set $ M = (B + B^2)/2 $. The encoder is constructed with a 256-neuron hidden layer and a 16-neuron embedding layer for all datasets.

%For our method, we set the clustering coefficient $\gamma$ to 1. The encoder is constructed with a 32-neuron hidden layer and a 16-neuron embedding layer for all benchmark datasets. We pretrain the autoencoder part for 100 iterations and then the whole model for 50 iterations. Target distribution $P$ is updated every 5 iterations. For Cora and Citeseer, we optimize the model with Adagrad algorithm and the learning rate are set to be 0.2. For the Pubmed dataset, we turns to an Adam optimizer with 0.01 learning rate to speed up the convergence.

% Table generated by Excel2LaTeX from sheet 'Cora'

\subsection{Experiment Results}

%We compare our DAEGC with baselines mentioned above on graph clustering first. Then we perform detailed analysis on coefficients in the model.
\vspace{.1cm}
% \subsubsection{Clustering Performance Comparison:}
The experiment results on the three benchmark datasets are summarized in Table \ref{tab:result_Cora}, \ref{tab:result_Citeseer}, and \ref{tab:result_Pubmed}, where the bold values indicate the best performance. C, S, and C\&S indicate if the algorithm uses only content, structure, or both content and structure information, respectively.  We can see that our method clearly outperforms all the baselines across most of the evaluation metrics.

% Table generated by Excel2LaTeX from sheet 'Pubmed'

\begin{figure}
\centering
\subfigure[ACC]
{
\label{fig:acc_embsz}
\includegraphics[width=0.47\columnwidth]{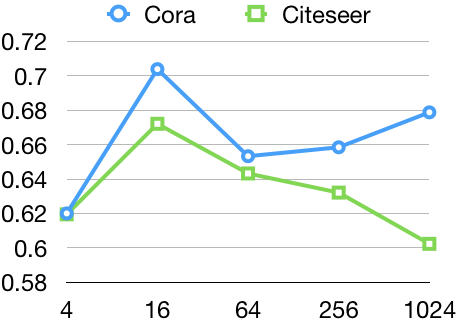}
}
\subfigure[NMI]
{
\label{fig:nmi_embsz}
\includegraphics[width=0.47\columnwidth]{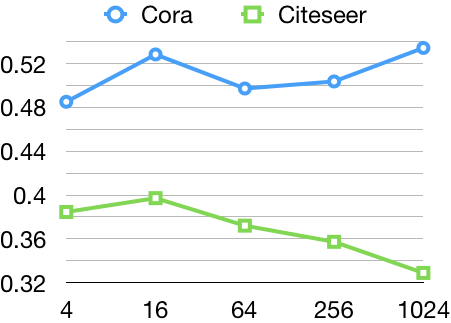}
}
\vspace{-3mm}
\caption{ Parameter v.s. different dimensions of the embedding}
\vspace{-3mm}
\label{fig:embsz}
\end{figure}

We can  observe from these results that methods using both the structure and content information of the graph generally perform better than those using only one side of information. In the Cora dataset, for example, TADW, GAE, VGAE and our method outperform all the baselines using one side of information. This observation demonstrates that both the graph structure and node content contain useful information for graph clustering, and illustrates the significance of capturing the interplay between two-sides information.

The results of most of the deep learning models are satisfactory. The GraphEncoder and DNGR algorithm are not necessarily an improvement although they both employ deep autoencoder for representation learning. This observation may result from their neglect at the node content information.

It is worth mentioning that our algorithm significantly outperforms GAE and VGAE. On the Cora dataset for example, our method represents a relative increase of 18.97\% and 29.49\% w.r.t. accuracy and NMI against VGAE, and the increase is even greater on the Citeseer dataset. The reasons for this are that 
%We can conclude that the observed superior performance of our proposed method benefit from 
(1) we employ a graph attention network that effectively integrates both content and structure information of the graph;
% (2) we use a deep architecture to learn the representation, which captures more underlying information; 
(2) Our self-training clustering component is specialized and powerful in improving the clustering efficiency.

\begin{table}[t]
  \centering
  \setlength\tabcolsep{4.2pt}
\begin{small}
  
  \vspace{-2mm}
    \begin{tabular}{c|c|cccc}
\toprule
        & Info. & ACC(\textcolor{red}{$\uparrow $})   & NMI(\textcolor{red}{$\uparrow $})    & F-score(\textcolor{red}{$\uparrow $})   & ARI(\textcolor{red}{$\uparrow $})  \\
\midrule
    $K$-means & C & 0.580 & \textbf{0.278} & 0.544 & 0.246 \\
    Spectral & S & 0.496 & 0.147 & 0.471 & 0.098 \\
    %Big-Clam & Structure & 0.394 & 0.001 & 0.510 & 0.000 \\
    GraphEncoder & S & 0.531 & 0.210 & 0.506 & 0.184 \\
    % DeepWalk & Structure & 0.6628 & 0.2561 & 0.5390 & 0.5315 & 0.5553 & 1.1421 & 0.2722 \\
    DeepWalk & S & 0.647 & 0.238 & 0.530 & 0.255 \\
    DNGR  & S & 0.468 & 0.153 & 0.445 & 0.059 \\
    M-NMF & S & 0.470 & 0.084 & 0.443 & 0.058 \\
\midrule
    %Circles\tnote{1} & Both  & - & - & - &-\\
    %RTM   & Both  & 0.575 & 0.194 & 0.444 & 0.149 \\
    RMSC  & C\&S  & 0.629 & 0.273 & 0.521 & 0.247 \\
    TADW  & C\&S  & 0.565 & 0.224 & 0.481 & 0.177 \\
    GAE  & C\&S  & 0.632 & 0.249 & 0.511 & 0.246 \\
    VGAE  & C\&S  & 0.619 & 0.216 & 0.478 & 0.201 \\
    DAEGC & C\&S  & \textbf{0.671} & {0.266} & \textbf{0.659} & \textbf{0.278} \\    
    %ep:92-10, seed:3, gamma:10, lr:0.01-0,002, 256-16
\bottomrule
    \end{tabular}%
    \vspace{-1mm}
    \caption{\normalsize Experimental Results on \textbf{Pubmed} Dataset}
  \label{tab:result_Pubmed}%
%   	\begin{tablenotes}
% 		\footnotesize
% 		\item[1] The Pubmed dataset is too large for the Circles algorithm.
% 	\end{tablenotes}
   \end{small}
  \vspace{-2mm}
\end{table}%

\vspace{.1cm}
\noindent\textbf{Parameter Study:}
We vary the dimension of embedding from 4 neurons to 1024 and report the results in Fig \ref{fig:embsz}. It can be observed from both \ref{fig:acc_embsz} and \ref{fig:nmi_embsz} that: when adding the dimension of embedding from 4-neuron to 16-neuron, the performance on clustering steadily rises; but when we further increase the neurons of the embedding layer, the performance fluctuates, though the ACC and NMI score both remain good on the whole.

\vspace{.1cm}
\noindent\textbf{Network Visulization:}
We visualize the Cora dataset in a two-dimensional space by applying the t-SNE algorithm \cite{van2014accelerating} on the learned embedding during training. The result in Fig \ref{fig:tsne} demonstrates that, after training with our graph attentional autoencoder, the embedding is already meaningful. However by applying self-training clustering, the embedding becomes more evident as our training progresses, with less overlapping and each group of nodes gradually gathered together.

\vspace{-2mm}
\section{Conclusion}
In this paper, we propose an unsupervised deep attentional embedding algorithm, DAEGC, to jointly perform graph clustering and learn graph embedding in a unified framework. The learned graph embedding integrates both the structure and content information and is specialized for clustering tasks. While the graph clustering task is naturally unsupervised, we propose a self-training clustering component that generates soft labels from ``confident" assignments to supervise the embedding updating. The clustering loss and autoencoder reconstruction loss are jointly optimized to simultaneously obtain both graph embedding and graph clustering result. A comparison of the experimental results with various state-of-the-art algorithms validate DAEGC's graph clustering performance.

\balance

\bibliographystyle{ijcai19}
\bibliography{DAEGC}

\end{document}